\documentclass{article}
\usepackage[nonatbib, preprint]{neurips_2022}
\usepackage[utf8]{inputenc} % allow utf-8 input
\usepackage[T1]{fontenc}    % use 8-bit T1 fonts
\usepackage{hyperref}       % hyperlinks
\usepackage{url}            % simple URL typesetting
\usepackage{booktabs}       % professional-quality tables
\usepackage{amsfonts}       % blackboard math symbols
\usepackage{nicefrac}       % compact symbols for 1/2, etc.
\usepackage{microtype}      % microtypography
\usepackage{xcolor}         % colors
\usepackage{graphicx}
\usepackage{amsmath}
\usepackage{authblk}
\usepackage{subfigure}
\usepackage[ruled]{algorithm2e}
\usepackage{multirow}

\hypersetup{
	colorlinks=true,
	linkcolor=red,
	filecolor=red,      
	urlcolor=red,
	citecolor=blue,
}

\title{Incremental Prototype Tuning for Class Incremental Learning}
\author[1, 2]{Jieren Deng}
\author[2]{Haojian Zhang}
\author[2]{Jianhua Hu}
\author[2]{Yunkuan Wang \thanks{Corresponding author} \space }
\affil[1]{School of Artificial Intelligence, University of Chinese Academy of Science}
\affil[2]{Institute of Automation, Chinese Academy of Sciences (CAS) \authorcr \{dengjieren2019, jianhua.hu, zhanghaojian2014, yunkuan.wang\}@ia.ac.cn}

\begin{document}

\maketitle

\begin{abstract}
  Class incremental learning(CIL) has attracted much attention, but most existing related works focus on fine-tuning the entire representation model, which inevitably results in much catastrophic forgetting. In the contrast, with a semantic-rich pre-trained representation model, parameter-additional-tuning (PAT) only changes very few parameters to learn new visual concepts. Recent studies have proved that PAT-based CIL can naturally avoid fighting against forgetting by replaying or distilling like most of the existing methods. However, we find that PAT-based CIL still faces serious semantic drift, the high-level forgetting problem caused by classifier learning bias at different learning phases, which significantly reduces the performance of PAT-based CIL. To address this problem, we propose  Incremental Prototype Tuning (IPT), a simple but effective method that tunes category prototypes for classification and learning example prototypes to compensate for semantic drift. Extensive experiments demonstrate that our method can effectively compensate for semantic drift. Combined with well-pre-trained Vit backbones and other PAT methods, IPT surpasses the state-of-the-art baselines on mainstream incremental learning benchmarks.

\end{abstract}

\section{Introduction} \label{section:introduction}
Open learning ability is the key to improve the level of artificial intelligence. Class incremental learning (CIL) \cite{rebuffi2017icarl} is exactly the natural way to raise the openness of AI systems, but it faces a notorious problem called catastrophic forgetting \cite{MCCLOSKEY1989109} -- directly end-to-end training the neural network learned from the old task on the new task leads to a sharp decline in the performance of the old task.

Most class incremental approaches resist catastrophic forgetting by replaying the seen data \cite{rebuffi2017icarl,liu2020mnemonics,xin2021memory} or distilling the previous models \cite{li2017learning,hou2019learning,douillard2020podnet}. However, when just feed a few new visual concepts, the representation model learned from last phase is still needed to be fine-tuned in the next phase endlessly, affecting the whole body of the model and bringing in lots of forgetting, so that it limits the performance of class incremental learning. 

Different from the fine-tuning methods that tune the entire model, parameter-additional-tuning methods (PAT) such as linear probing, prompt tuning, and adapter, only allow a very small number of additional parameters to be trained. Therefore PAT can effectively weaken forgetting while retaining the ability to learn new knowledge, as well as reducing the overfitting of each small task in CIL with the support of rich prior knowledge stored in the pre-trained model. In addition, PAT converges quickly because of its little learnable parameter scale. A recent study L2P\cite{l2p} has proved that PAT with a strong pre-trained representation model can even help class incremental learning to obtain competitive results without fine-tuning the entire representation model, replaying old data, or distilling from the previous model. We believe that PAT is a very promising method to apply CIL to practical applications.

\begin{figure*}[htbp]
  \centering
  \includegraphics[width=12cm]{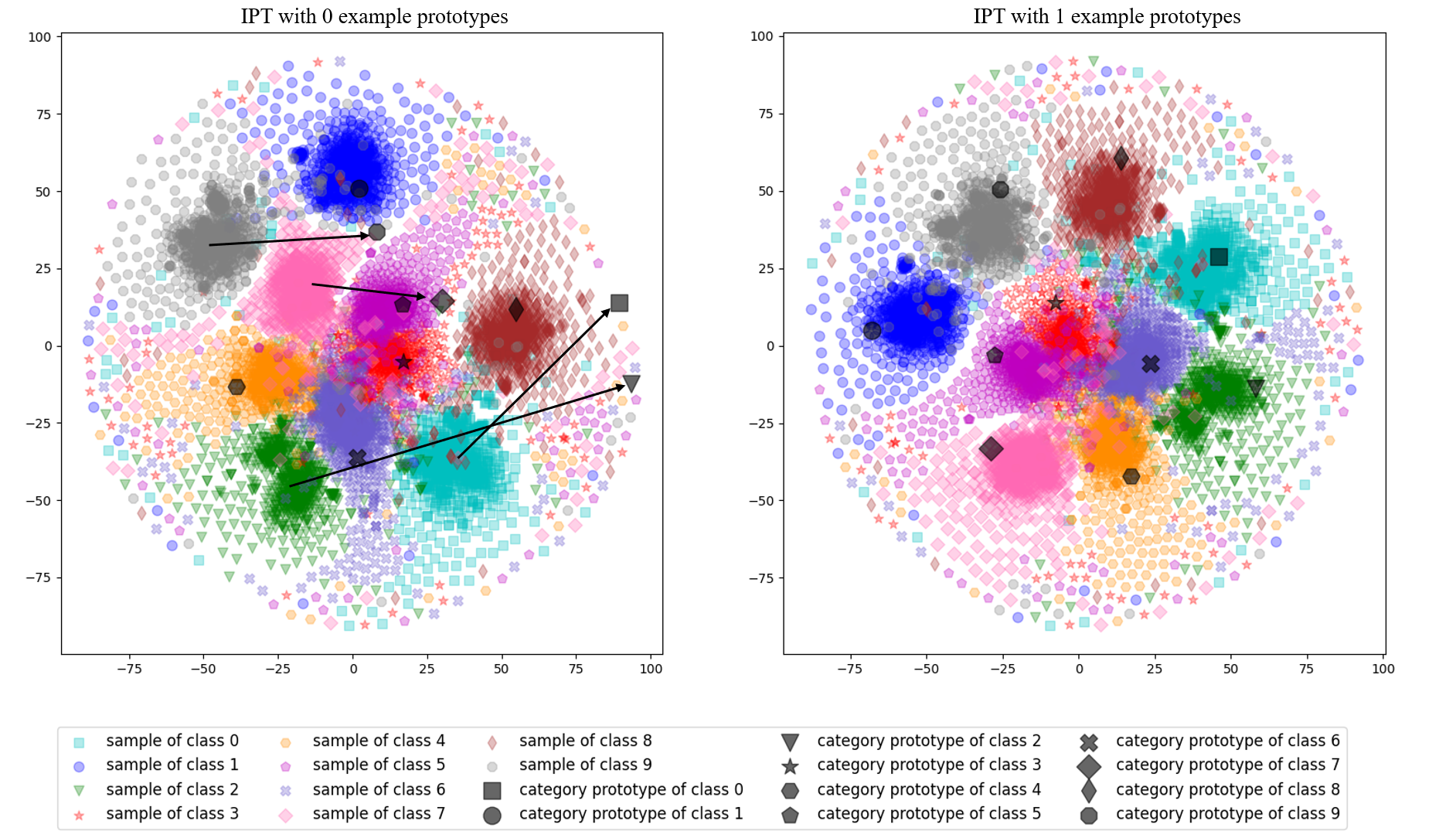}
  \caption{{\bf T-SNE\cite{van2008visualizing} of CIFAR-10 \cite{krizhevsky2009learning} to show semantic drift and its compensation}. Note that the black arrows reflect the semantic drift. IPT without example prototypes may cause serious semantic drift as shown in the left picture, in which the category prototypes may be far away from the density center of the target class and even close to another class. Only one example prototype can even greatly alleviates this problem as shown in the right picture.}
  \label{fig:semantic_drift}
  \end{figure*}

Although PAT with strong pre-trained representation can greatly improve the performance of class incremental learning systems while avoiding replaying original samples, the performance is not satisfactory enough. We find a major reason is that the categories at different phases are not trained together, and the classifier may overlap the category prototypes learned at different phases in the representation space, as shown in the left subfigure of Figure \ref{fig:semantic_drift}, which is called semantic drift \cite{yu2020semantic}.

To address semantic drift, we propose a novel PAT method. In detail, extract several learnable prototypes for each category (namely example prototypes). These example prototypes are classified by all learned category prototypes to constrain the classifier to isolate the category prototypes when learning new categories from different phases. Example prototypes aim to imitate each class' distribution in representation space, so they should not be homogenized. In order to ensure the diversity of example prototypes, we initialize them by K-means clustering and design a new maximum similarity loss to optimize them. Summarizing the above steps, our method called incremental prototype tuning (IPT) can effectively reduce semantic drift as shown in the right subfigure of Figure \ref{fig:semantic_drift}. Unlike replaying methods storing the original samples, IPT only stores several prototypes of each category thus saving a lot of storage resources.

By combining IPT and L2P, our method can outperform baselines and others state-of-the-art models on CIFAR-100 \cite{krizhevsky2009learning}, ImageNet-100 \cite{rebuffi2017icarl} and ImageNet-1000 \cite{russakovsky2015imagenet} benchmarks. IPT is indeed simple but surprisingly enables an AI system to learn new concepts incrementally without replaying original data or distillation from previous models.

In summary, we provide the following contributions:

1. We argue that PAT is a promising direction for CIL, and point out that semantic drift is one of the major factor that affects the performance of PAT-based CIL. To quantitatively analyze this effect, we also propose a new metric called category drift rate (CDR) to measure semantic drift.

2. We propose a novel PAT-based method named incremental prototype tuning (IPT), which alleviates semantic drift by classifying the learnable example prototypes. A new maximum similarity loss (MSL) is designed to maintain the diversity of example prototypes to ensure the effectiveness of the classification constraints of the example prototypes.

3. IPT achieves the new state-of-the-art on standard incremental learning benchmarks without replaying original data and distilling from previous models, which reduces the requirement of memory storage and improves the feasibility of PAT-based CIL in practical application.
  
\section{Related work}
\subsection{Class Incremental Learning}
Deep learning models overspecialize at individual tasks while lacking the ability to learn openly to deal with other tasks, which is in sharp contrast to human beings. To address this problem, incremental learning attempts to develop an artificial intelligence system that can continuously learn and process new tasks from new data, while retaining the knowledge learned from previous tasks. Incremental learning can be divided into three categories \cite{van2019three}: task incremental learning, domain incremental learning, and class incremental learning (CIL) \cite{rebuffi2017icarl}, where CIL is more challenging and closer to the needs of practical applications which the task identity is not informed when classifying test samples. Recent popular CIL methods can be categorized into three classes: replay-based, regularization-based, and parameter-isolation-based \cite{de2019continual,prabhu2020gdumb,liu2021adaptive}. Replay-based methods preserve a small amount of data in previous tasks in memory or a generative model in order to replay these data when training the model on new data to overcome catastrophic forgetting \cite{rebuffi2017icarl, liu2020mnemonics, xin2021memory, iscen2020memory, welling2009herding}. The regularization-based methods provide terms in the final loss to restrict the model with prior or knowledge distillation to change too much to forget previous knowledge \cite{kirkpatrick2017overcoming, lee2017overcoming, zeng2019continual, li2017learning, hou2019learning, douillard2020podnet}. For parameter-isolation-based methods, different subsets of the model parameters are dedicated to each task to prevent the previous tasks from getting any possible forgetting \cite{mallya2018packnet, serra2018overcoming, rusu2016progressive, aljundi2017expert}.

\subsection{Parameter-additional-tuning with pre-trained model}
Parameter-additional-tuning (PAT) aims to alias the pre-trained representations and features of downstream tasks by adjusting the additional task-specific parameters while remaining the pre-trained model fixed. Recently, the most popular PAT can be categorized into two kinds of tuning methods: adapter-based, and prompt-based. The adapter-based approach injects small-scale neural network modules (adapters) connected to the transformer layer, and only adjusts these adapters for model adaptation \cite{houlsby2019parameter, NEURIPS2021_081be9fd, ruckle2021adapterdrop}. The adapter module contains a down-projection layer and an up-projection layer and achieves impressive performance while only owning about 0.5\%~8\% parameters of the whole model. Instead of injecting neural modules to the Transformer model, the prompt-based approach learns new input tokens for transformer backbone \cite{gao2021making, li-liang-2021-prefix, hu2022knowledgeable}, has achieved promising performance in various NLP tasks. L2P \cite{l2p} is the first work that introduces the idea of prompt-tuning into incremental learning. We choose to use it as one of our baseline methods.
  
\section{Incremental prototype tuning}
\subsection{Problem setup}
Given a data stream of a labeled sample super set $ \{X^{1},X^{2},\cdots,X^{t}\} $, where $ X^y=\{x_{1}^{y},x_{2}^{y},\cdots,x_{n_{y}}^{y}\} $ is a set containing all samples belonging to class $ y $, class incremental learning models learn from a dataset stream \{$ D_{0},D_{1},\cdots,D_{t},\cdots $\}. At each phase t, only $ D_t=\{X^{s_t+1},X^{s_t+2},\cdots,X^{s_{t+1}}\} $ is accessible for training the model, while the seen data $ {X^1,\cdots,X^{s_{t}} } $ is no longer available. Here, $ s_t $ represents the sum of the number of categories learned before phase t. During the test, the trained model is expected to classify all seen categories. It is needed to note that joint training whose accuracy is regarded as the upper bound of class incremental learning puts all samples of all categories together for training.

\subsection{Baseline}
\begin{figure}[htbp]
  \centering
  \includegraphics[width=6cm]{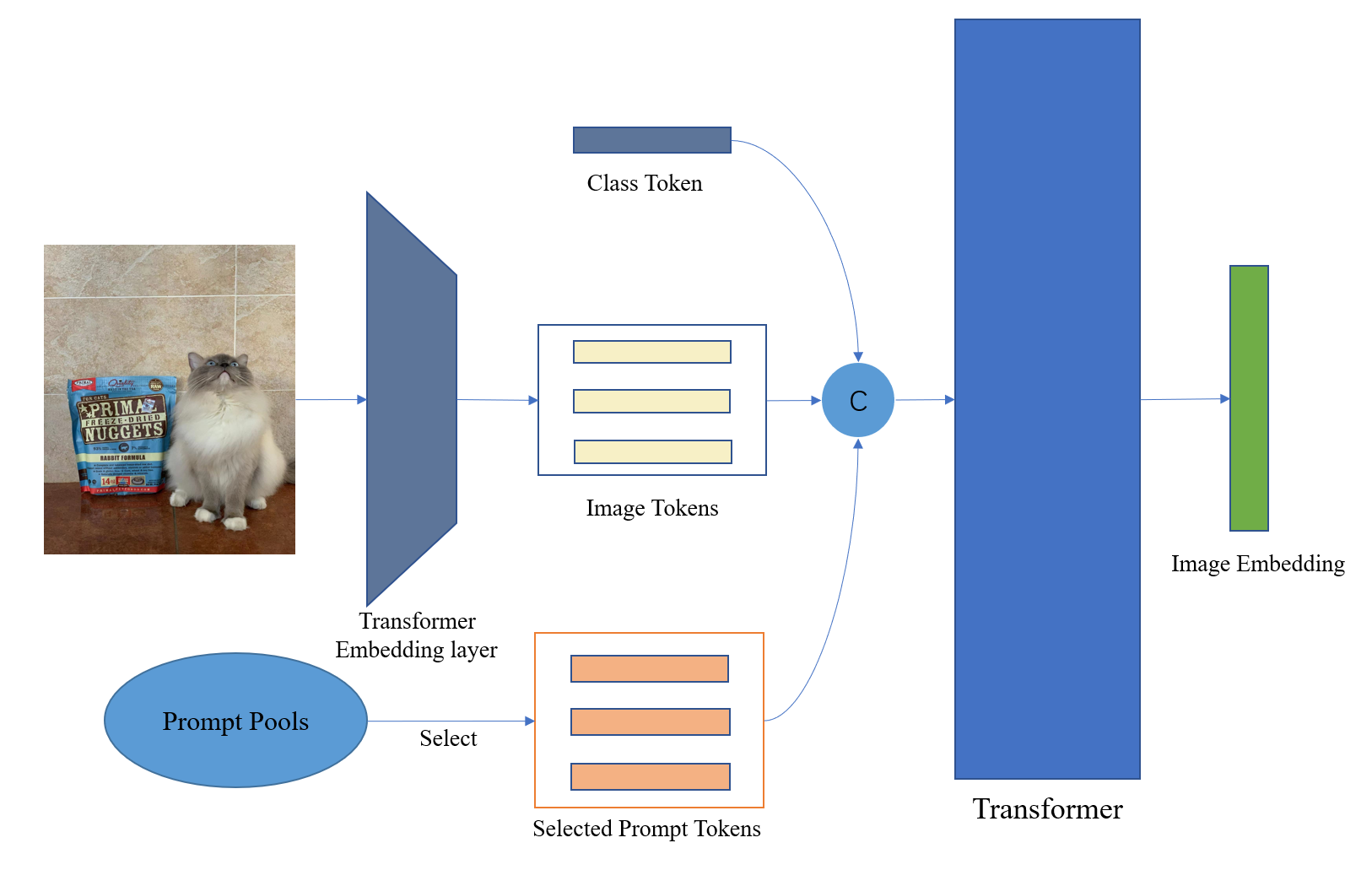}
  \caption{\textbf{Structure of L2P.}``C" means concatenation.}
\label{fig:l2p}
\end{figure}

L2P is one of the baseline methods of our work. As shown in Figure \ref{fig:l2p}. It learns a pool of prompt queries as one part of the input for the pre-trained Vit \cite{dosovitskiy2021image} backbone and mitigates catastrophic forgetting by a key-value-pair-based prompt selection strategy. Though both IPT and L2P are PAT methods, the difference between IPT and L2P is that IPT chooses to learn category prototypes as the input for the classifier and servers a plug-in module for all PAT methods. Since semantic drift is a higher-level forgetting problem, it is less obvious to see the semantic drift when the representation model is not frozen because it is mixed with a lower level of forgetting (representation forgetting). Therefore, another baseline of our work is the method that naively freezes the backbone as the representation model to avoid the interference of representation forgetting.

\subsection{Overview}
\begin{figure*}[htbp]
  \centering
  \includegraphics[width=10cm]{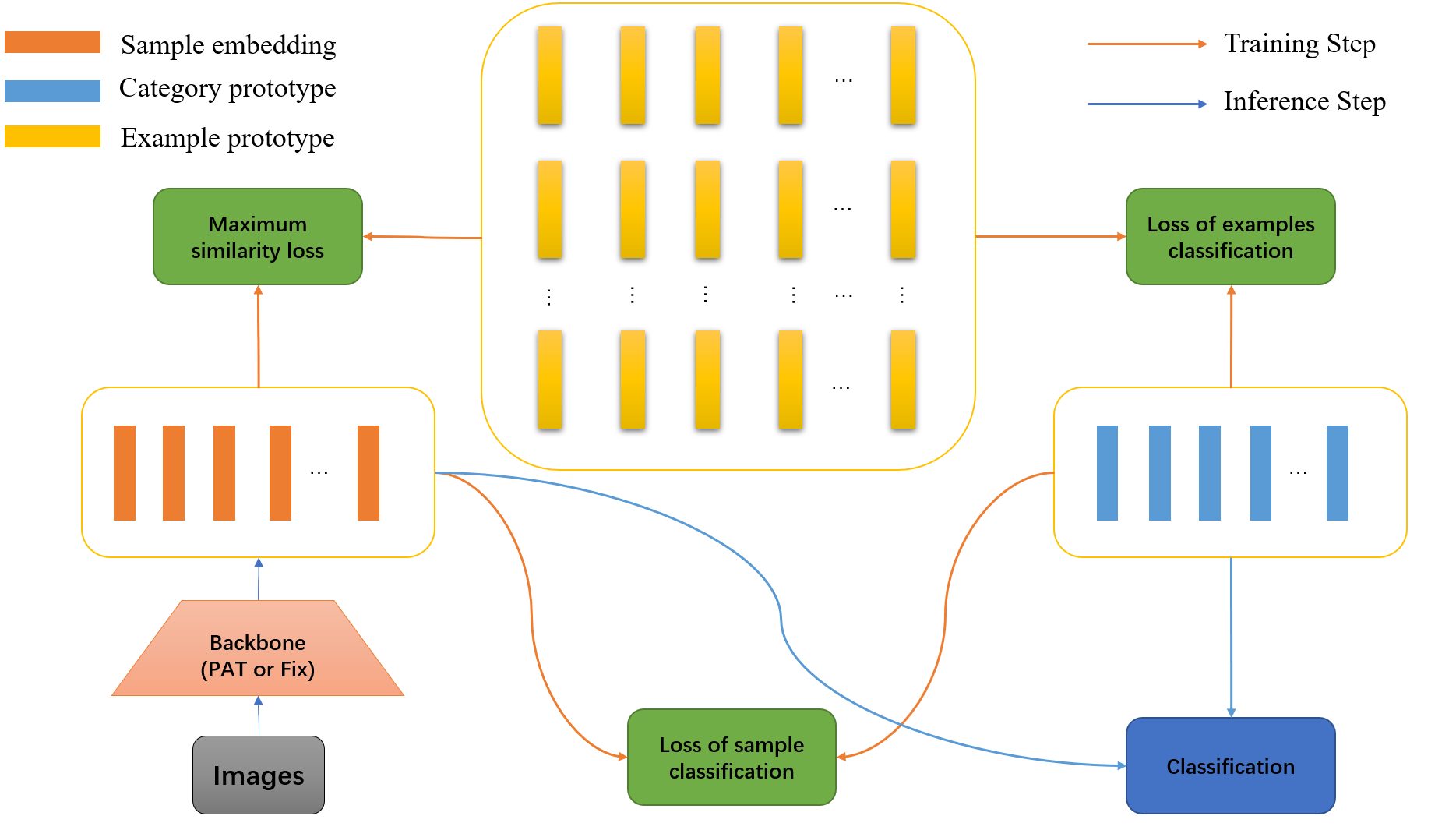}
  \caption{{\bf Overview of IPT.} IPT consists of category prototypes and example prototypes that are optimized by three kinds of loss including maximum similar loss, loss of sample classification, and loss of example classification. The backbone can be frozen or tuned by any PAT method such L2P.}
\label{fig:overview}
\end{figure*}

An overview of incremental prototype tuning (IPT) for class incremental learning is presented in Figure \ref{fig:overview}. The IPT consists of a pre-trained model with additional parameters, category prototypes, and examples prototypes.

\textbf{Three Stage Training strategy.} The training strategy of our model for each phase involves three stages, the initialization stage, the prototype training stage, and the additional parameter tuning stage. 

At the initialization stage, new category prototypes and new examples prototypes are initialized and integrated into the model  The new category prototypes can simply be initialized as the sample embeddings of target categories or the embeddings of the class name provided by the text encoder pre-trained together with the representation model, such as CLIP \cite{radford2021learning}. The new example prototypes are initialized by a clustering algorithm. 

At the prototype training stage, the loss at phase t is computed by the simple addition:
\begin{align}
    L^t=L_{CES}^{t}+L_{CEP}^{t}+L_{MS}^{t}. 
\end{align}
At this stage, only category prototypes, example prototype of phase t are trainable.

At the additional parameter tuning stage, we add the parameter-additional-tuning loss term $L_{PAT}^{t}$ to the previous loss:
\begin{align}
    L^t=L_{PAT}^{t}+L_{CES}^{t}+L_{CEP}^{t}+L_{MS}^{t}.
\end{align}
At this stage, only additional parameters are trainable. If there is no additional parameter for tuning, the training of this stage can be canceled.
  
\textbf{Inference.} The inference architecture consists of the pre-trained model, additional parameters, and all category prototypes. Samples are classified according to which category prototypes their embeddings, extracted by the pre-trained representation model with additional parameters, are most similar with.
  
\subsection{Loss of sample classification}
At the prototype training stage during phase $t$, each sample is extracted as a representation $\Phi(x, \theta)$ by the pre-trained representation model $\theta$. For each sample x, suppose the similarity between $\Phi(x, \theta)$ and category prototype $c_i$ is represented as:
\begin{align}
  s(x, c_i) = <\Phi(x, \theta), c_i>,
\end{align}
where $ i \in \{s_t,\cdots,s_{t+1} - 1 \} $ and $ x \in D_{t} .$
We get the classification result of samples after applying a softmax function:
\begin{align}
  \begin{split}
    [y_{s_t}(x),\cdots,y_{s_{t+1}-1}(x)]^T = softmax([s(x,c_{s_t}) \\,\cdots,s(x,c_{s_{t+1}-1})]^T).
  \end{split}
\end{align}
Finally, cross-entropy loss during phase t is computed as:

\begin{align} \label{eq:lces}
  L_{CES}^t(x) = \sum_{i=s_t}^{s_{t+1}-1}\hat y_i(x)\log{y_i(x)},
\end{align}
where $\hat y_i(x)$ is the ground truth of probability of sample $x$ belongs to class $i$.

At the additional parameters tuning stage, the parameter-additional-tuning loss $L_{PAT}^{t}$ is computed by the similar computing way of $L_{CES}^t$, the minor difference is that each sample is extracted as a representation $\Phi(x, \theta, \gamma)$ by the pre-trained representation model $\theta$ with additional parameters $\gamma$. Sometimes $L_{PAT}^{t}$ may also contains an additional loss such as prompt key select loss in L2P \cite{l2p}, which is dependent by which parameter-additional-tuning method is used.

\subsection{Learning example prototypes}\label{section:sec3.3}
\textbf{Motivation.} As shown in Figure \ref{fig:semantic_drift}, when we map category prototypes and sample embeddings into low dimension space for visualization, we find that category prototypes are not always located in the distribution of the target class sample embeddings, instead, it may be located in the distribution of other class embeddings, which is called semantic drift \cite{yu2020semantic}. Since the samples at the former and latter phases are not available simultaneously, as long as the category prototype can correctly classify the samples at the current phase, the loss function does not necessarily continue to encourage the category prototypes to enter the sample representation density center of the target class. The category prototypes may even occupy the sample representation density center of the categories at other phase, resulting in classification confusion that greatly harms the performance of PAT. So that we propose to optimizing the classification loss of several example prototypes to prevent semantic drift.
  
\textbf{Maximum similarity loss (MSL).} We define the example prototypes of class $ i $ as $ e_i^j, i=1,\cdots,N_c,j=1,\cdots,N_e$, where $ N_c $ is the total number of all classes, and $ N_e $ is the number of example prototypes. Example prototypes are supposed to imitate the distribution of representation embeddings of every category, they are trained by the maximum similarity loss (MSL) computed as:
\begin{align} \label{eq:lms}
  L_{MS}^t(x) = 1 - \max \limits_{j} (<\Phi(x, \theta), e^j_{g(x)}>),
\end{align}
where $ j = 1, 2, \cdots, N_e$, $ x \in D_t$, and $g(x)$ is ground truth class id label of sample $ x $. Note that we compute $\Phi(x, \theta)$ without additional parameters.Maximum similarity does not require each sample prototype to be close to each sample of the target category. In this way, it is possible to find some difficult samples that deviate from the density center of the sample.
  
\textbf{Initialization of example prototypes.} The optimization direction of example prototypes is mainly dominated by maximum similarity loss. When example prototypes are far from the target category embedding center at the initialization stage, only a few example prototypes can eventually enter the sample distribution space of the target category, resulting in the example prototype's underfitting of target class distribution. Therefore, the example prototype needs a better initialization method to shorten the initial distance between all example prototypes and the target category embedding center. Note that this distance should not be too small, otherwise example prototype will lose the ability to mine difficult samples. So we use k-means as our initialization method. For each new class $ C $, we extract the representation of all its training samples, then use the k-means algorithm to split them into $ N_e $ clusters, and then use these cluster centers as the initialization values of new example prototypes. This initialization method not only allows example prototypes to cover the sample representation distribution of the corresponding class, but also retains the diversity of example prototypes.

\subsection{Loss of example classification} 
Semantic drift can be prevented by classification loss of example prototypes. During phase $ t $, classification loss of example prototypes is computed as:
\begin{align} \label{eq:lcep}
  L^t_{CEP}(e_i^j) = - \sum_{k=1}^{s_{t+1}-1} \hat{y_k}(e_i^j)\log{y_k(e_i^j)},
\end{align}
where $ i \in \{1, 2, \cdots, s_{t+1} - 1\}$,
\begin{align}
  [y_k(e_i^j)]^T = softmax([<e_i^j, c_k>]^T), 
\end{align}
where $  k \in \{1, 2, \cdots, s_{t+1} - 1\}$.
  
\textbf{Discussion.} Classification loss of example prototypes forces $ c_k $ to be far away from $ e_i^j $ for each $i \neq k$, which helps to isolate category prototypes from different phases. Moreover, since it can also adjust those category prototypes born before phase $ t $, accuracy of phase $t^b (t^b < t)$  can even  be continually improved, achieving positive backward transfer. Another advantage of example prototypes is that they naturally avoid the class-imbalanced problem, since the number of example prototypes used in equation \ref{eq:lcep} for each category is equal which makes $L^t_{CEP}$ a class-balanced loss.
  
\section{Experiment}
\subsection{Experiment setup} \label{section:experiment_setup}
\textbf{Benchmark rotocols.} We validate our method with two widely used benchmark protocols including the protocol of half initialization (PHI) proposed and the protocol of no initialization (PNI). The PHI starts training the models on half the classes (i.e., $ 50 $ for CIFAR-100 and ImageNet-100, $ 500 $ for ImageNet-1000), then equally divides the remaining classes and incrementally learns them at the future phase. The PNI simply divides the total classes equally among phases and learns them phase by phase. Note that an $ n $ phases PHI has actually $ n+1 $ phases with one phase for half initialization, while PNI has exactly $n$ phases.
  
\textbf{Metrics.} We use four indicators to evaluate our method: average incremental accuracy, final accuracy, forgetting rate, and CDR. Rebuffi et al. \cite{rebuffi2017icarl} introduce average incremental accuracy (AIA), which averages the accuracy of all phases. Forgetting rate (FR), proposed by Liu et al. \cite{liu2020mnemonics}, measures the degree of performance degradation in the first initialization phase after incremental training. Besides, in order to see more clearly the influence of semantic drift, we propose a new metric named category drift rate (CDR) to measure semantic drift as followed:
\begin{align}
  CDR = N_d / N_c,
\end{align}
where $N_d$ is the number of categories that misclassify more than 50 \% of all the samples that are classified as that category, and $N_c$ is the total number of categories. The CDR is computed after all training phased on the test dataset.
% whose distance to the sample feature center of the target category is farther than that to the sample feature center of any other category and 

\begin{figure*}[htbp]
  \centering
  \includegraphics[width=14cm]{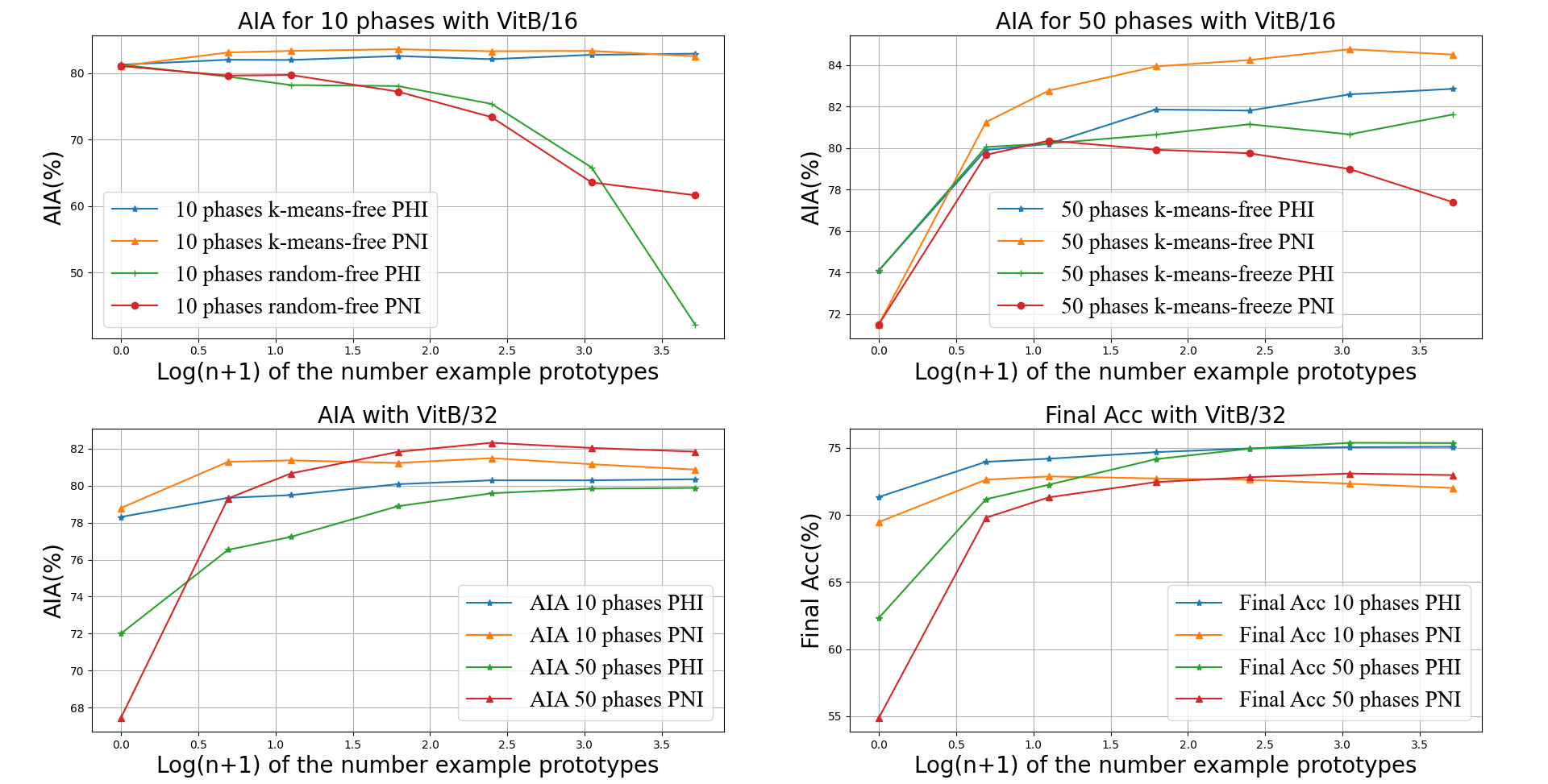}
  \caption{Relationship curves between IPT-based class incremental learning performance and the number of example prototypes. In order to avoid infinity and display better, the horizontal axis adopts the logarithm of the number of example prototypes plus one. The legend content ``k-means'' or ``random'' means whether to use k-means to initialize the example prototypes and "freeze" or "free" means whether to freeze those category prototypes before the current learning phase.}
  \label{fig:prototypes}
\end{figure*}

\textbf{Implementation details.} We implement our method in PyTorch with an RTX 3090. We train 20 epochs at the prototype training stage and 10 epochs at the additional parameters tuning stage with batch size 32 for all datasets. SGD is used for optimization with a base learning rate of 0.01. Momentum and weight decay parameters are set to 0.9 and 0.0005, respectively. We multiply the learning rate by 0.1 at the beginning of the 5th, 10th, and 15th epoch. We use a clipping gradient in the range of 5 to 100 to accelerate training. We choose cosine distance to uniformly measure the distance among sample embeddings, category prototypes, and example prototypes. Code will be available after review.

\subsection{Main results}
\begin{table*}[htbp]
  \centering
  \resizebox{\textwidth}{!}{
    \begin{tabular}{c|c|ccc|ccc|cc}
    \hline
    \multirow{2}[0]{*}{Metric} & \multirow{2}[0]{*}{Method} & \multicolumn{3}{c|}{CIFAR100} & \multicolumn{3}{c|}{Standford Car} & \multicolumn{2}{c}{Oxford 102 Flower} \\
          &       & phase=5 & 10    & 25    & 7     & 14    & 49    & 3     & 17 \\
    \hline
    \multirow{4}[0]{*}{AIA(\%) $\uparrow$ } & Fixed ViTB/32 & 78.05  & 75.61  & 68.37  & 72.97  & 69.37  & 60.64  & 93.69  & 88.29  \\
          & Fixed ViTB/32 + IPT & 80.54  & 80.08  & 79.22  & 80.81  & 80.23  & 80.01  & 93.74  & 94.42  \\
          & L2P ViTB/16 & 79.56  & 76.72  & 73.88  & 78.35  & 76.60  & 67.43  & 93.33  & 78.23  \\
          & L2P ViTB/16 + IPT & 81.47  & 79.73  & 79.42  & 83.34  & 82.41  & 83.32  & 92.21  & 93.20  \\
    \hline
    \multirow{4}[0]{*}{Final Acc(\%) $\uparrow$ } & Fixed ViTB/32 & 72.14  & 68.28  & 57.96  & 63.25  & 56.22  & 43.19  & 90.19  & 81.40  \\
          & Fixed ViTB/32 + IPT & 75.45  & 75.11  & 74.75  & 75.76  & 75.14  & 74.18  & 90.96  & 92.39  \\
          & L2P ViTB/16 & 74.24  & 70.93  & 65.33  & 68.59  & 65.78  & 52.41  & 88.86  & 57.20  \\
          & L2P ViTB/16 + IPT & 76.70  & 74.64  & 74.36  & 78.08  & 76.68  & 76.51  & 88.81  & 90.10  \\
    \hline
    \multirow{4}[0]{*}{CDR(\%) $\downarrow$ } & Fixed ViTB/32 & 7  & 12  & 26  & 17  & 22  & 37  & 2  & 6  \\
          & Fixed ViTB/32 + IPT & 7  & 6  & 7  & 4  & 6  & 8  & 1  & 1  \\
          & L2P ViTB/16 & 7  & 11  & 20  & 19  & 20  & 32  & 3  & 34  \\
          & L2P ViTB/16 + IPT & 6  & 7  & 7  & 3  & 4  & 4  & 2  & 1  \\
    \hline
    \multirow{2}[0]{*}{Upper Bound} & Fixed ViTB/32 & \multicolumn{3}{c|}{77.89} & \multicolumn{3}{c|}{80.39} & \multicolumn{2}{c}{94.96} \\
          & L2P ViTB/16 & \multicolumn{3}{c|}{83.91} & \multicolumn{3}{c|}{83.18} & \multicolumn{2}{c}{96.01} \\
    \hline
    \end{tabular}%
  }
    \caption{Comparison with baselines on CIFAR100, Standford Car, and Oxford 102 Flower under PHI protocol.}
  \label{tab:main_results}%
\end{table*}%

\textbf{Comparison with baselines.} We evaluate the proposed IPT on CIFAR100 \cite{krizhevsky2009learning}, Standford-Cars \cite{KrauseStarkDengFei-Fei_3DRR2013}, and Oxford-102-Flower \cite{Nilsback08}. We use two baseline methods including L2P \cite{l2p} and a naive fixed backbone with category prototypes for comparison. The models are pre-trained on CLIP400M \cite{radford2021learning}, a dataset consists of $ 400 $ million image-text pairs. The main results of comparison with the baseline methods is shown in Table \ref{tab:main_results}. An intuitive impression of these results is that high CDR will lead to low accuracy, while IPT can significantly reduce CDR and improve accuracy, which verifies the hypothesis that semantic drift seriously reduces the performance of PAT-based CIL and the effectiveness of IPT for mitigating semantic drift.
% Table \ref{tab:main_results} shows the main results of comparison with the baseline methods, which illustrates that our method can consistently improve AIA and the final accuracy of the baselines by significantly mitigating the semantic drift measured by CDR.

\begin{table*}[htbp]
  \begin{center}
    \resizebox{\textwidth}{!}{
    \begin{tabular}{c|cccc|ccc}
    \hline
    Protocol & \multicolumn{4}{c|}{PHI} & \multicolumn{3}{c}{PNI} \\
    \hline
    Phases & \multicolumn{2}{c}{5} & \multicolumn{2}{c|}{10}& 5  & 10  & 50  \\
    \hline
    Metrics & AIA(\%)$\uparrow$ & FR(\%)$\downarrow$ & AIA(\%)$\uparrow$ & FR(\%)$\downarrow$ & AIA(\%)$\uparrow$ & AIA(\%)$\uparrow$ & AIA(\%)$\uparrow$ \\
    \hline
    iCaRL-CLIP-VitB/16 & 58.18 ± 0.57 & 31.88 & 52.78 ± 0.96 & 33.1 & 71.45 ± 0.44 & 66.01 ± 0.92 & 57.08 ± 0.84 \\

    UCIR-CLIP-VitB/16 & 62.98 ± 1.02 & 18.3 & 60.56 ± 0.71 & 22.34 & 63.77 ± 0.84 & 58.65 ± 0.73 & 56.76 ± 3.54 \\

    BiC-CLIP-VitB/16 & 57.86 ± 0.36 & 31.22 & 53.28 ± 0.91 & 31.5 & 73.30 ± 0.45 & 68.89 ± 1.10 & 62.19 ± 0.75\\

    DyTox-CLIP-VitB/16 & 76 ± 1.12 & 6.77  & 75.82 ± 1.21 & 8.30  & 77.42 ± 0.62 & 77.03 ± 1.24 & 75.12 ± 1.13 \\

    % DER & 72.60 ± 0.78 & \_  & 72.45 ± 0.76 & \_ & 75.55 ± 0.65 & 74.64 ± 0.28 & 72.05 ± 0.55 \\
    \hline
    L2P-CLIP-VitB/16 & 79.56 & 4.86  & 76.72 & 6.42 & 83.42 & 80.17 & 70.20 \\
    L2P-CLIP-VitL/14 & 86.90 & 2.86  & 85.54 & 2.48 & 88.60 & 84.28 & 77.80 \\
    \hline
    Fixed CLIP-VitB/16 + IPT & 82.32 ± 0.86 &  0.047 &   82.1 ± 1.07 &  0.187 &  83.12 ± 2.00 & 83.30 ± 2.40 & 84.25 ± 2.40 \\
    Fixed CLIP-VitL/14 + IPT & \textbf{87.73 ± 0.55} & \textbf{-0.4} & \textbf{87.45 ± 0.83} & \textbf{-0.46} & \textbf{88.62 ± 1.22} & \textbf{87.83 ± 1.35} & \textbf{89.09 ± 1.98} \\
    \hline
    \end{tabular}%
    }
  \caption{{Comparison with state-of-the-art on CIFAR-100 benchmarks}}
  \label{tab:sota_cifar100}%
\end{center}
\end{table*}%

\begin{table*}[h]
  \begin{center}
  \resizebox{\textwidth}{!}{
  \begin{tabular}{c|cccc|cccc}
    \hline
    {Datasets} & \multicolumn{4}{c|}{ImageNet-100} & \multicolumn{4}{c}{ImageNet-1000} \\
    \hline
    {Protocols} & \multicolumn{4}{c|}{PHI} & \multicolumn{4}{c}{PHI} \\
    \hline
    {Phases} & \multicolumn{2}{c}{5} & \multicolumn{2}{c|}{10} & \multicolumn{2}{c}{5} & \multicolumn{2}{c}{10} \\
    \hline
    {Metrics} & {AIA(\%)}$\uparrow$ & {FR(\%)}$\downarrow$ & {AIA(\%)$\uparrow$} & {FR(\%)}$\downarrow$ & {AIA(\%)$\uparrow$} & {FR(\%)}$\downarrow$ & {AIA(\%)$\uparrow$} & {FR(\%)}$\downarrow$\\
    \hline
    iCaRL-CLIP-VitB/16 & 66.56 & 42.4  & 62.0  & 42.36 & 53.34 & 25.03 & 42.78 & 31.56 \\
    UCIR-CLIP-VitB/16 & 72.15 & 30.12 & 68.82 & 31.56 & 66.45 & 22.18 & 62.34 & 25.41 \\
    BiC-CLIP-VitB/16 & 70.98 & 25.14 & 66.34 & 30.14 & 43.45 & 29.14 & 42.29 & 33.71 \\
    DyTox-CLIP-VitB/16 & 78.53 & 7.56 & 76.44 & 8.34 & 71.80 & 4.30 & 69.54 & 6.37 \\
    % DER & \_ & \_ & 77.73 & \_ & \_ & \_ & \_ & \_ \\
    \hline
    L2P-CLIP-VitB/16 & 76.63 & 0.44 & 75.92 & 2.44 & 77.13 & 1.41 & 75.51 & 2.09 \\
    L2P-CLIP-VitL/14 & 75.62 & 0.88  & 71.49 & 2.6 & 79.64 & 0.37 & 75.31 & 3.42 \\
    \hline
    Fixed CLIP-VitB/16 + IPT & 82.75 &  0.56  &  83.22 & 0.73 & 79.43 & \textbf{-0.90} & 79.12 & \textbf{-0.74} \\
    Fixed CLIP-VitL/14 + IPT & \textbf{85.64} & \textbf{0.16}  & \textbf{85.98} & \textbf{ 0.04} & \textbf{83.37} & -0.41 & \textbf{83.10} & \ 0.31 \\
    \hline
\end{tabular}%
}
\caption{Comparison with state-of-the-art on ImageNet-100 and ImageNet-1000 benchmarks}
\label{tab:sota_imageNet}%
\end{center}
\end{table*}%

% % Table generated by Excel2LaTeX from sheet 'Sheet1'
% \begin{table*}[htbp]
%   \centering
%   % \resizebox{\textwidth}{!}{
%     \begin{tabular}{c|ccccccc}
%     \hline
%     Number of Example Prototype & 0     & 1     & 2     & 5     & 10    & 20    & 40 \\
%     Extra Memory Proportion (EMP) & EMP(\%) & EMP(\%) & EMP(\%) & EMP(\%) & EMP(\%) & EMP(\%) & EMP(\%) \\
%     \hline
%     Class Number = 0 & 0     & 0     & 0     & 0     & 0     & 0     & 0 \\
%     Class Number = 1 & 0.00\% & 0.00\% & 0.00\% & 0.01\% & 0.01\% & 0.03\% & 0.05\% \\
%     Class Number = 10 & 0.01\% & 0.01\% & 0.04\% & 0.08\% & 0.15\% & 0.28\% & 0.54\% \\
%     Class Number = 100 & 0.13\% & 0.13\% & 0.40\% & 0.79\% & 1.45\% & 2.77\% & 5.41\% \\
%     \hline
%     \multicolumn{4}{c|}{Entire Model Size(MB)} & \multicolumn{4}{c}{614.2} \\
%     \hline
%     \end{tabular}%
%   % }
%     \caption{Extra memory requirements of different number of example prototypes under PNI settings. Extra memory proportion refers to the proportion of additional parameter memory in the total model memory. All the result in this table are based on a fixed VitB/16 on CIFAR-100.}
%     \label{tab:memory}%
% \end{table*}%

\begin{table}[htbp]
  \centering
    \begin{tabular}{c|ccc}
    \hline
    Number of class & 1     & 10    & 100 \\
    \hline
    NE = 0 & 0.00\% & 0.01\% & 0.13\% \\
    NE = 1 & 0.00\% & 0.03\% & 0.26\% \\
    NE = 5 & 0.01\% & 0.08\% & 0.79\% \\
    NE = 10 & 0.01\% & 0.15\% & 1.45\% \\
    NE = 40 & 0.05\% & 0.54\% & 5.41\% \\
    \hline
    \multicolumn{2}{c|}{Entire Model Size(MB)} & \multicolumn{2}{c}{614.2} \\
    \hline
    \end{tabular}%
    \caption{\textbf{Extra memory proportion} of different number of example prototypes under PNI settings. Extra memory proportion refers to the proportion of additional parameter memory in the total model memory. ``NE" means the number of example prototypes. All the results in this table are based on a fixed VitB/32 on CIFAR-100.}
  \label{tab:memory}%
\end{table}%

\textbf{Comparison with state-of-the-art.} We also compare our methods with the state-of-the-art results of several non-PAT-based methods including DyTox \cite{douillard2021dytox}, BiC \cite{Wu_2019_CVPR}, UCIR \cite{hou2019learning}, iCaRL \cite{rebuffi2017icarl} and the recent state-of-the-art PAT-based method L2P \cite{l2p}. All the results are based on the models including Vit-B/16, Vit-L/14 \cite{dosovitskiy2021image} pre-trained by CLIP \cite{radford2021learning} on CLIP400M. $ 10 $ example prototypes are used for IPT. We run experiments for different protocols and phase settings on CIFAR100, ImageNet100 \cite{rebuffi2017icarl} and ImageNet1000 \cite{russakovsky2015imagenet}. As shown in Table \ref{tab:sota_cifar100} and \ref{tab:sota_imageNet}, our method performs better than other methods on different incremental phase settings, and the average incremental accuracy does not decrease significantly along the increase of phases.  In particular, the FR of our method is very close to zero and even negative, suggesting that the model has achieved positive backward transfer. On the contrary, due to too many parameters being fine-tuned, the non-PAT-based methods are easy to cause overfitting on split incremental tasks and suffer from huge forgetting, making themself unable to take advantage of large-scale pre-trained models.
 
\subsection{Ablation of prototypes.} 
\textbf{Settings.} We conduct a further study about the contribution of example prototypes and category prototypes for our method. In detail, we employ different numbers of example prototypes (including number zero which means there is no example prototype used) under the PHI and PNI setting for both 10 and 50 phases on CIFAR-100. We also compare random and k-means initialization for the example prototypes. The results of whether to freeze those category prototypes born before phase $n$ while learning at phase $n+t$ to prevent those knowledge learned in new phases to be transferred to old phases are also submitted. VitB/16 and VitB/32 pre-trained on CLIP400M are used as the fixed representation models in this section.

\textbf{Results.} Figure \ref{fig:prototypes} exhibits the result of the prototype ablation study. The methods of using example prototypes are better than the methods of not using, particularly under the more challenging settings such as the 50-phase setting. Although one example prototype is enough to achieve obvious improvement under 10 phase setting, the performances of IPT-based methods under more difficult settings still need more example prototypes to be ensured. However, there is an obvious marginal effect in the number of used example prototypes. Also, the more example prototypes are used, the more IPT depends on initialization methods, and of course, the more storage space is required. It is also worth noting that the methods that do not freeze the categories prototypes born before phase $n$ while learning at phase $n+t$ outperform those methods that freeze. In other words, a positive backward transfer is actually achieved behind the tuning of category prototypes.

\subsection{Further study}
\textbf{Extra memory cost.} IPP only adds category prototypes and example prototypes to the model at each phase. Since all the additional prototypes are vectors only, the incremental memory requirement rises slowly, as shown in Table \ref{tab:memory}. The concern that using too many example prototypes will bring a large memory footprint can be addressed by just setting the number of example prototypes to $ 1 $. As shown in Figure \ref{fig:prototypes}, one example prototype can successfully solve semantic drift while only maintaining the same memory consumption as the classifier weight (category prototypes).

\textbf{Ablation of maximum similarity loss.} To see the effectiveness of the maximum similarity loss (MSL) of our method, we also replace the equation \ref{eq:lms} by adopting  average similarity loss (ASL) as the followed equation:
\begin{align} 
  L_{AS}^t(x) = 1 -  \frac{\sum\limits_{j}(<\Phi(x, \theta), e^j_{g(x)}>)}{N_e}.
\end{align}
By comparing the performances of MSL and ASL, Table \ref{tab:similarity_loss} demonstrates that IPT does benefit from MSL. The more phases the incremental learning split the dataset into, the more effective MSL is.

\begin{table}[htbp]
  \centering
  \resizebox{8cm}{!}{
    \begin{tabular}{c|cc|cc}
    \hline
    Setting & \multicolumn{2}{c|}{10 Phase PNI} & \multicolumn{2}{c}{50 Phase PNI} \\
    Metrics & AIA(\%)$\uparrow$ & Final Acc(\%)$\uparrow$ & AIA(\%)$\uparrow$ & Final Acc(\%)$\uparrow$ \\
    \hline
    MSL   & 82.07  & 73.59  & 82.19 & 73.49 \\
    ASL   & 80.14  & 72.00  & 78.46 & 68.21 \\
    \hline
    \end{tabular}%
  }
  \caption{\textbf{Ablation results of maximum similarity loss}. All the results are produced based on a frozen VitB/16 pre-trained CLIP400M.Ten example prototypes are used for this experiment.}
  \label{tab:similarity_loss}%
\end{table}%

\section{Conclusions} \label{section:con}
In this work, we emphasize that semantic drift is a serious problem that limits the performance of PAT-based CIL, and propose IPT to alleviate semantic drift. At each new phase t, IPT freezes the examples prototypes born before phase t, generates new category prototypes and examples prototypes, and optimize them with two classification losses and maximum similarity loss. Exhaustive experiments show that our method can significantly reduce semantic drift and improve the performances of PAT-based CIL methods without replaying original samples and distilling from previous models. The superior performance of our method further verifies that PAT-based CIL is promising.

% \textbf{Potential negative societal impact.} IPT can be applied to many downstream applications. However, it may still have a potential societal impact, since IPT relies on models pre-trained on large-scale datasets that may have bias and ethical problems. These issues may be carried out in the process of IPT incremental learning. Therefore, we encourage any users to thoughtfully check the datasets for pre-training to exclude any bias and fairness issues.

IPT reduces the high-level forgetting (semantic drift) of the PAT-based CIL method but still remains low-level forgetting (representation forgetting) unsolved.  Although fixing the representation can prevent representation forgetting, it hinders learning better feature extraction ability. This is an open problem about balancing plasticity and stability.

% In additional, IPT still needs memory that grows linearly with the number of categories. Future works will focus on how to balance plasticity and stability of the PAT-based method, while limiting the memory footprint to a small and fixed amount.

% Use \bibliography{yourbibfile} instead or the References section will not appear in your paper
{
\small
\bibliographystyle{splncs04}
\bibliography{my_egbib}
}
% Use \bibliography{yourbibfile} instead or the References section will not appear in your paper
\end{document}